\documentclass[10pt,twocolumn,letterpaper]{article}

\usepackage{authblk}
\usepackage[pagenumbers]{iccv} 
\usepackage{multirow}
\usepackage{algorithm}
\usepackage{algorithmic}
\usepackage{textcomp}
\usepackage{listings}

\definecolor{iccvblue}{rgb}{0.21,0.49,0.74}
\usepackage[pagebackref,breaklinks,colorlinks,allcolors=iccvblue]{hyperref}

\def\paperID{5644} 
\def\confName{ICCV}
\def\confYear{2025}

\title{OutDreamer: Video Outpainting with a Diffusion Transformer}

\author[1]{Linhao Zhong\thanks{Contribute Equally.}$^{,}$}
\author[2]{Fan Li$^{*,}$\thanks{Corresponding Author.}$^{,}$}
\author[3]{Yi Huang}
\author[3]{Jianzhuang Liu}
\author[2]{Renjing Pei}
\author[2]{Fenglong Song}
\affil[1]{Shanghai Jiao Tong University}
\affil[2]{Huawei Noah’s Ark Lab}
\affil[3]{Shenzhen Institute of Advanced Technology, Chinese Academy of Sciences}

\date{} 

\usepackage{orcidlink}
\usepackage{enumitem}
\newcounter{suppfigure}

\newcommand{\suppfigref}[1]{\ref{#1}}
\newcounter{supptable}

\newcommand{\supptabref}[1]{\ref{#1}}

\begin{document}
\maketitle
\begin{abstract}

    Video outpainting is a challenging task that generates new video content by extending beyond the boundaries of an original input video, requiring both temporal and spatial consistency. Many state-of-the-art methods utilize latent diffusion models with U-Net backbones but still struggle to achieve high quality and adaptability in generated content. Diffusion transformers (DiTs) have emerged as a promising alternative because of their superior performance. We introduce OutDreamer, a DiT-based video outpainting framework comprising two main components: an efficient video control branch and a conditional outpainting branch. The efficient video control branch effectively extracts masked video information, while the conditional outpainting branch generates missing content based on these extracted conditions. Additionally, we propose a mask-driven self-attention layer that dynamically integrates the given mask information, further enhancing the model's adaptability to outpainting tasks. Furthermore, we introduce a latent alignment loss to maintain overall consistency both within and between frames. For long video outpainting, we employ a cross-video-clip refiner to iteratively generate missing content, ensuring temporal consistency across video clips. 
    Extensive evaluations demonstrate that our zero-shot OutDreamer outperforms state-of-the-art zero-shot methods on widely recognized benchmarks.
    
\end{abstract}  

\section{Introduction}

\begin{figure*}[ht]
    \centering
    \includegraphics[width=\linewidth]{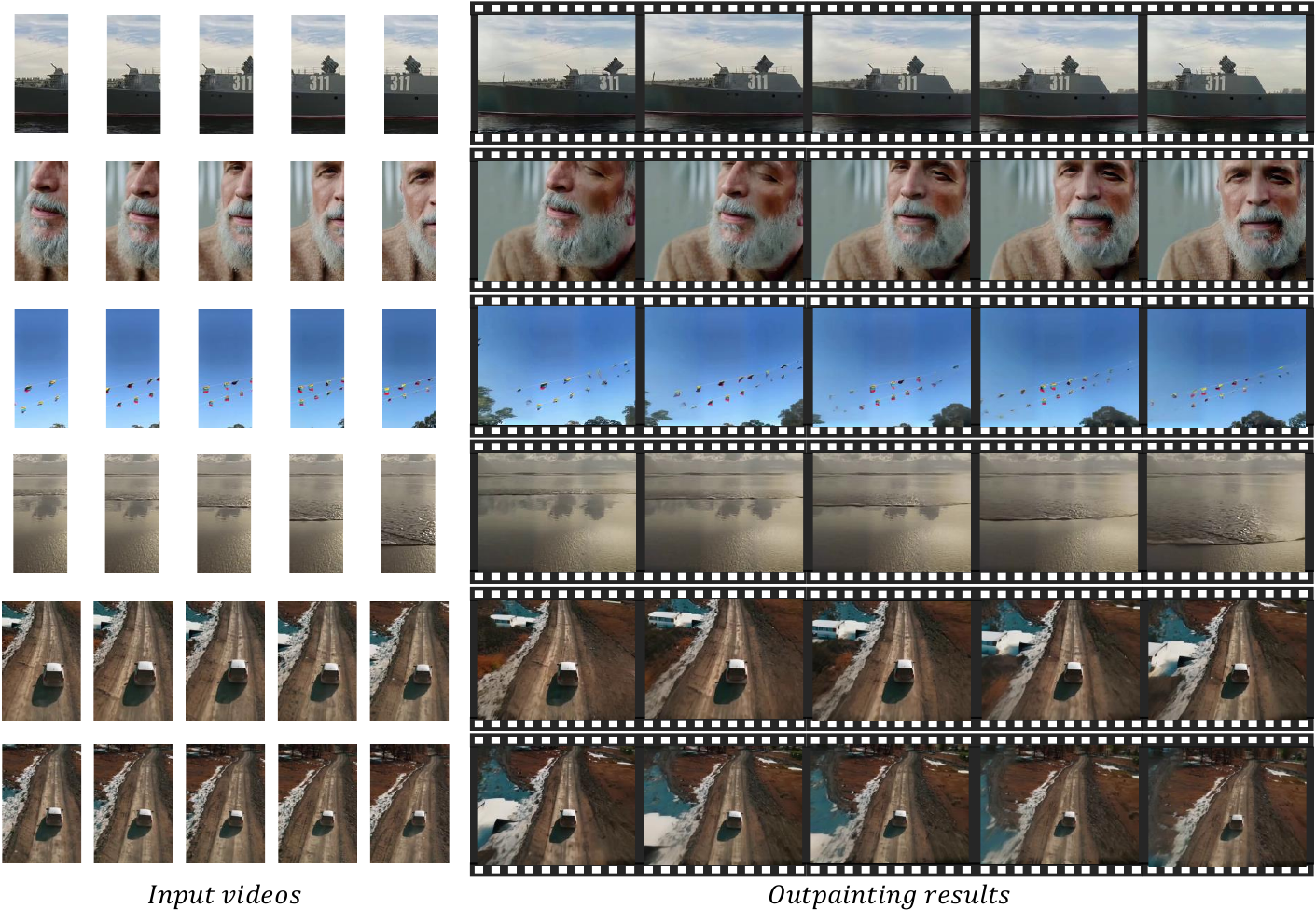}
    \caption{OutDreamer is a zero-shot video outpainting method based on the DiT backbone, producing high-quality missing content efficiently. It can generate long videos iteratively and maintain inter-clip consistency.}
    \label{example}
\end{figure*}

Video outpainting, the task of generating new content beyond the boundaries of an original input video, is a challenging problem that requires both spatial and temporal consistency. In recent years, this task has gained significant attention. While image outpainting~\cite{image-outpaint-Maskgit, image-outpaint-Inout, image-outpaint-edge-guided, image-outpaint-Palette, image-outpaint-sketch-guided, image-outpaint-semantics-guided}, which generates missing content outside the image boundaries, has been widely studied and achieved remarkable results, video outpainting is inherently more complex due to the additional challenge of maintaining consistency across frames. The generated content must not only appear visually realistic but also remain temporally coherent.

Current approaches to video outpainting can be broadly categorized into two groups: zero-shot methods and one-shot learning methods. 
Zero-shot methods generate missing content directly from the input video without any fine-tuning on the specific video. Dehan~\cite{Dehan} and M3DDM~\cite{M3DDM} are representative zero-shot methods. 
Dehan uses background estimation and optical flow to ensure temporal consistency. 
M3DDM employs a masked 3D diffusion model based on a U-Net backbone, utilizing bidirectional learning with mask modeling to generate missing content. 
However, the videos produced by current zero-shot methods often lack visual realism, with unstable results and blurry outpainting regions.
The second group, one-shot learning methods, includes approaches like MOTIA~\cite{Be-your-outpainter}. These methods require fine-tuning on each specific input video before generating the missing content. 
Although effective, this approach is time-consuming, often producing repetitive content in the outpainting regions.


In this work, we aim to develop a more general and efficient zero-shot video outpainting method that maintains high-quality outputs. To this end, we propose OutDreamer, a novel approach for video outpainting. 
OutDreamer consists of two primary components: an efficient video control branch for extracting essential video conditions and a conditional outpainting branch for generating missing content based on these conditions along with iterative generation for long videos. 
Unlike traditional diffusion models with U-Net backbones, OutDreamer is based on a diffusion transformer (DiT)~\cite{DiTs} for its ability to capture long-range dependencies in video data.



In order to better utilize the network capabilities of the DiT architecture and the performance of the pre-trained model, we aim to inject masked video condition at the early stages of the network. ControlNeXt~\cite{peng2024controlnext} demonstrates the effectiveness of injecting condition right after the first block of DiTs for video generation, and we adopt a similar injection strategy in this work. 
Additionally, we propose a mask-driven self-attention layer to replace the standard self-attention layer in each DiT block. This layer dynamically integrates the given mask information through several light-weight linear layers and activation functions, allowing each block to freely allocate attention between masked and unmasked regions.
Furthermore, we introduce a novel latent alignment loss function in the latent space to mitigate common inconsistency issues between the generated and original parts of the video in outpainting.
For long video outpainting, we propose an iterative method and a cross-video-clip refiner to maintain inter-clip consistency. 
Figure~\ref{example} shows some results generated by OutDreamer.


Our main contributions are summarized as follows:
\begin{itemize} 
    \item \textbf{DiT-Based Video Outpainting Framework}: We propose a framework with two main components: a control branch and a conditional outpainting branch. The control branch effectively extracts masked video information, while the conditional outpainting branch integrates this information into the denoising model at the early stages of the network. Additionally, we design a mask-driven self-attention layer that dynamically incorporates the given mask information into the self-attention process, improving outpainting quality.

    \item \textbf{Latent Alignment Loss}: We introduce a latent alignment loss to align the mean and variance of each frame, ensuring consistency across frames.

    \item \textbf{Iterative Long Video Outpainting}: We develop an iterative approach for generating long videos and a cross-video-clip refiner to maintain inter-clip consistency. By using the last few frames of one clip to guide the generation of the next, our method helps bridge gaps between clips and improves temporal consistency.

    \item \textbf{State-of-the-Art Performance}: We conduct extensive experiments to evaluate our OutDreamer and show that it outperforms existing zero-shot video outpainting methods. We also perform ablation studies to validate the effectiveness of each module.
\end{itemize}
\section{Related Work}

\subsection{Image Inpainting and Outpainting}
Image inpainting and outpainting, often considered sub-tasks of image editing, have distinct objectives and challenges~\cite{quan2024deep, huang2024diffusion, xiang2023deep}. Image inpainting involves filling in missing or occluded regions of an image with plausible and natural content. Traditional learning-based methods usually use CNNs or Transformers~\cite{li2018context, yang2017high, li2022srinpaintor, liu2023coordfill} to directly learn a mapping from input to output images. Unlike this deterministic approach, some earlier methods rely on Generative Adversarial Networks (GANs) to model the data distribution~\cite{li2024image, suvorov2022resolution, li2022mat, zhao2021large}. 
Recently, diffusion models have gained prominence in this area due to their capacity to capture complex data distributions and their stable training processes~\cite{li2024magiceraser, xie2023smartbrush, yang2023uni, manukyan2023hd}.

Image outpainting, though related to inpainting, specifically focuses on generating new pixels to seamlessly extend the boundaries of an image. InOut~\cite{image-outpaint-Inout} formulates this problem from the perspective of inverting GANs to find optimal latent codes. 
Recent text-to-image diffusion models, such as Stable Diffusion~\cite{rombach2022high} and DALL-E~\cite{ramesh2022hierarchical}, can also be adapted for outpainting, as they are trained on diverse datasets with varying image sizes and shapes. 

\subsection{Video Inpainting and Outpainting}
Extending image inpainting methods to the video domain would introduce unique challenges, especially in maintaining temporal consistency across frames. To address them, various approaches have been developed, including 3D CNNs~\cite{chang2019free, hu2020proposal}, video Transformers~\cite{liu2021fuseformer, zhang2022flow, zhou2023propainter}, and flow-guided propagation~\cite{zhang2022inertia, xu2019deep, gao2020flow}. 
However, the generated content often lacks sufficient detail. With the success of diffusion models in image generation and inpainting, recent efforts have explored the use of pre-trained text-to-image models for video inpainting~\cite{wang2023zero, zhang2024avid, qi2023fatezero, zi2024cococo}.
For instance, AVID~\cite{zhang2024avid} adopts a similar architecture to AnimateDiff~\cite{guo2023animatediff}, initializing the inpainting model with image-based modules and fine-tuning a motion module initialized from a pre-trained AnimateDiff model. 

As for video outpainting, previous approaches often address this task in a zero-shot manner. For example, Dehan~\cite{Dehan} decomposes the video outpainting process into five stages: flow estimation, background estimation, flow completion, video completion, and post-processing. M3DDM~\cite{M3DDM} trains a 3D diffusion model using a masked modeling strategy directly. However, they usually result in blurry or unrealistic content. Instead, MOTIA~\cite{Be-your-outpainter} fine-tunes each video individually, leveraging both the intrinsic patterns of the source video and the generative capabilities of pre-trained diffusion models, but it can be time-consuming and impractical for real-world applications. 

\section{Preliminaries}
\label{Sec:Preliminaries}

\noindent \textbf{Diffusion Formulations.}
Diffusion models~\cite{DDPM, IDDPM, DDIM} assume a forward noising process following the Markov chain, which gradually applies noise to a data sample $x_0$ from its real data distribution $q(x)$  and obtains a sequence of noisy samples $x_t$ in $T$ steps with a variance schedule $\beta_1,\ldots,\beta_T$: $q(x_{t}\vert x_{t-1}) = \mathcal{N}(x_{t};\sqrt{1-\beta_t}x_{t-1},\beta_t \mathbf{I})$. 
The closed form of the forward process can be expressed as $x_t = \sqrt{\bar{\alpha}_t}x_0 + \sqrt{1-\bar{\alpha}_t}\epsilon,$
where $\alpha_t=1-\beta_t$, $\bar{\alpha}_t=\prod_{i=1}^t \alpha_i$, and $\epsilon \sim \mathcal{N}(\mathbf{0}, \mathbf{I})$.

Diffusion models are trained to learn the reversed process through a joint distribution $p_{\theta}(x_{0:T})$ that follows the Markov chain with parameters $\theta$: $p_{\theta}(x_{t-1}\vert x_t) = \mathcal{N}(x_{t-1};\mu_{\theta}(x_t,t), \Sigma_{\theta}(x_t,t)).$
The parameters $\theta$ are usually optimized by a neural network $\epsilon_{\theta}(x_t,t)$ that directly predicts noise vectors $\epsilon_t$ instead of $\mu_{\theta}$ and $\Sigma_{\theta}$ with the following simplified objective~\cite{song2021score}:
\begin{equation}
	\mathbb{E}_{x_0,t,\epsilon_t\sim\mathcal{N}(\mathbf{0},\mathbf{I})}\Big[\vert\vert\epsilon_t - \epsilon_{\theta}(x_t,t)\vert\vert^2 \Big].
	\label{eq:training_obj}
\end{equation}

\noindent\textbf{Diffusion Transformers (DiTs).}
Latent diffusion models (LDMs)~\cite{rombach2022high} for image generation commonly use a U-Net backbone, while DiTs~\cite{DiTs} replace it with a transformer in the latent space. DiTs adhere to the best practices of Vision Transformers (ViTs)~\cite{ViTs}, which have demonstrated superior scalability for visual recognition compared to traditional convolutional networks.

\begin{figure*}[ht]
    \centering
    \includegraphics[width=\linewidth]{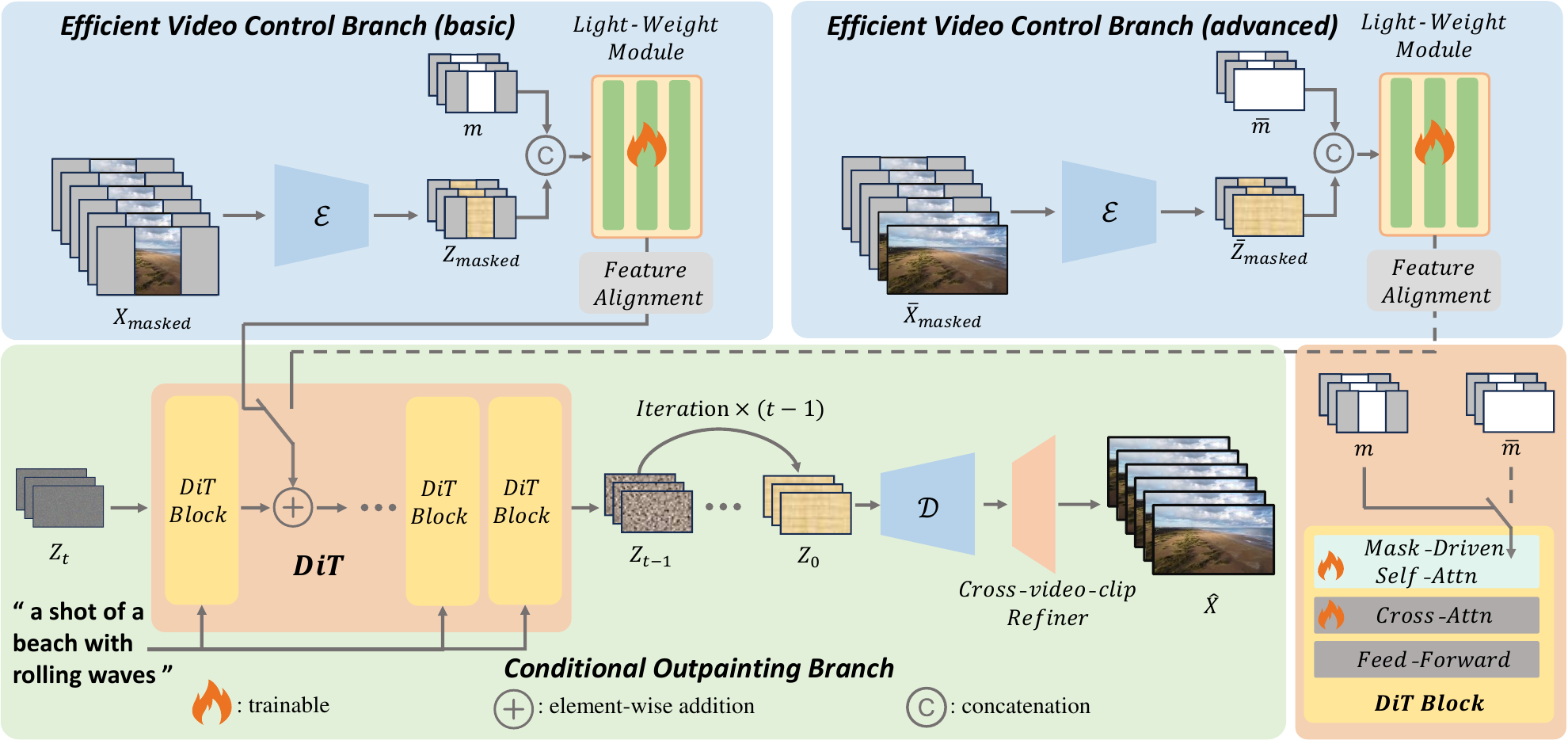}
    \caption{\textbf{Overview of OutDreamer.}
    OutDreamer consists of two main components: an efficient video control branch for extracting essential video conditions and a conditional outpainting branch for generating missing content based on these conditions and iterative generation for long videos. The efficient video control branch has two versions: a basic version shown in the upper-left part and an advanced version in the upper-right part. The lower-right part illustrates the DiT block with mask-driven self-attention. 
    }
    \label{overview}
\end{figure*}

\section{Method}
Given an input video $X' \in \mathbb{R}^{H' \times W' \times S \times 3}$ and its corresponding outpainting masks $M \in \mathbb{R}^{H \times W \times S \times 1}$, the video outpainting task aims to generate an extended video $\hat{X} \in \mathbb{R}^{H \times W \times S \times 3}$ based on the masked video condition $X_{\text{masked}} \in \mathbb{R}^{H \times W \times S \times 3}$, where $H' < H$, $W' < W$, and $X_{\text{masked}}$ is padded from $X'$.

\subsection{Overview of OutDreamer}
As shown in Figure~\ref{overview}, our OutDreamer, a DiT-based outpainting framework, consists of two main branches: an efficient video control branch and a conditional outpainting branch based on a DiT backbone. 
The efficient video control branch has two versions: a basic version shown in the upper-left part of Figure~\ref{overview}, and an advanced version in the upper-right part. The advanced version, combined with a cross-video-clip refiner in the conditional outpainting branch, is utilized for long video generation that will be described in Section~\ref{Sec:LongVideoGeneration}. For now, we focus solely on the basic version of the efficient video control branch.

Specifically, the spacetime VAE encoder $\mathcal E$ maps the masked video condition $X_{\text{masked}}$ into the latent space, producing latent masked video frames $Z_{\text{masked}} \in \mathbb{R}^{h \times w \times s \times d}$. Latent masks $m \in \mathbb{R}^{h \times w \times s \times 1}$ are obtained by downsampling the outpainting masks $M \in \mathbb{R}^{H \times W \times S \times 1}$.
OutDreamer then applies a light-weight module to extract features from the video condition $Z_{\text{masked}}$ and the masks $m$, aligns the extracted features, and integrates them into the DiT backbone, ultimately outputting the denoised latent representation. To make the DiT block more suitable for video outpainting and improve the generation quality, we replace the standard self-attention layer with a mask-driven self-attention layer, as shown in the lower-right part of Figure~\ref{overview}.
After several denoising steps, the latent representation of the generated video is obtained and mapped back to the pixel space using the VAE decoder $\mathcal D$. For long videos that cannot be generated in a single pass, we adopt an iterative generation approach and use a cross-video-clip refiner for inter-clip alignment.




\subsection{Efficient Video Control Branch}

Inspired by the emerging controllable generation approach ControlNeXt~\cite{peng2024controlnext}, OutDreamer employs a few 2D convolution layers $\mathcal{F}_c(\cdot ;\Theta_c)$ with learnable parameters $\Theta_c$, to extract features from $Z_{\text{masked}}$ and $m$. Subsequently, after alignment, these extracted features are added to the features outputted from the first block of the DiT backbone $\mathcal{F}_m(\cdot;\Theta_d)$ with pre-trained learnable parameters $\Theta_d$. Specifically, the integrated features $y_c$ are calculated as follows: 
\begin{equation}
    y_c = \mathcal{F}_m(Z_t; \Theta_{d'}) 
    + \eta(\mathcal{F}_c(Z_{\text{masked}}, m; \Theta_c); \mu_m, \sigma_m )
\end{equation}
where $\Theta_{d'}$ represents the parameters of the first block of the backbone, $\Theta_{d'} \subseteq \Theta_d$, $\Theta_c<<\Theta_d$, $Z_t\in\mathbb{R}^{h \times w \times s \times d}$ is the noisy video latent at time step $t$, and $\mu_m$ and $\sigma_m$ are the mean and variance of the denoising features after the first block of the DiT backbone. The feature alignment function $\eta$ aligns the conditional features with the conditional outpainting branch features in terms of mean and variance, normalizing them to ensure the stability and effectiveness of the training process.

\subsection{Mask-Driven Self-Attention}

\begin{figure}[t]
    \centering
    \includegraphics[width=\linewidth]{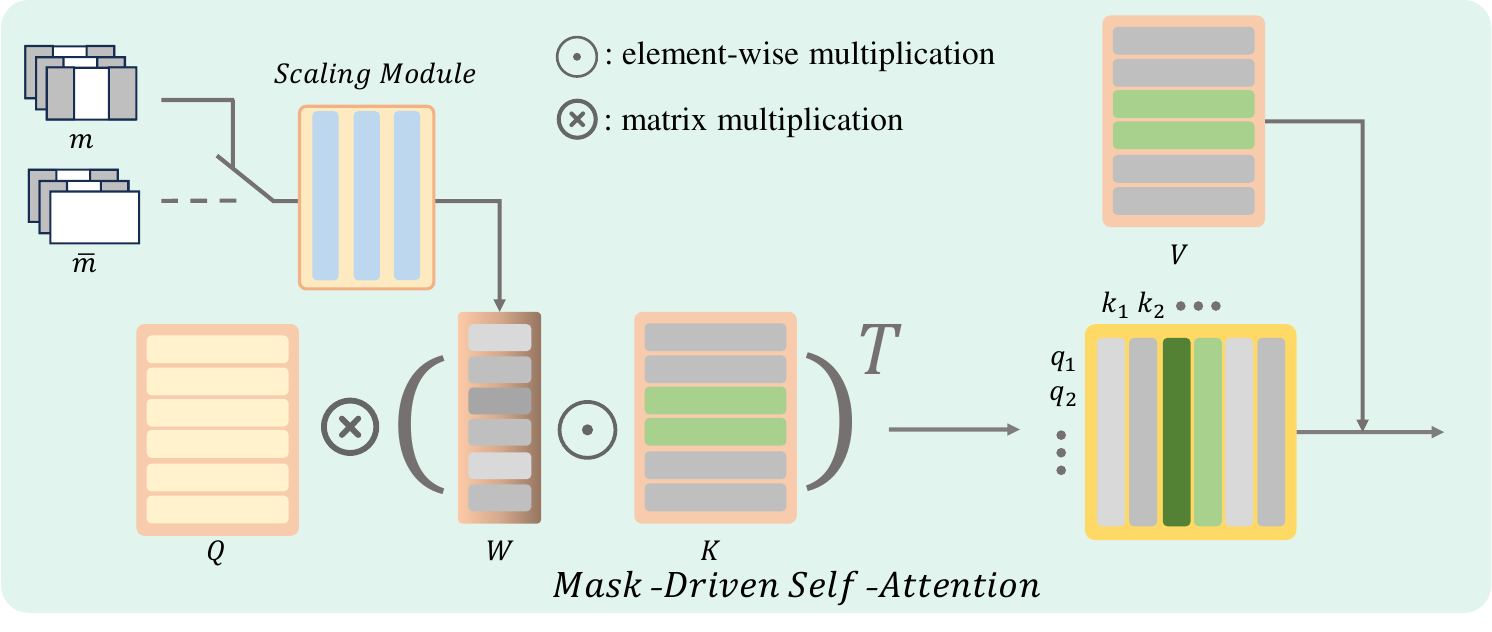}
    \caption{\textbf{Mask-Driven Self-Attention.} The green embeddings represent patches corresponding to the given region, while the gray embeddings represent patches corresponding to the outpainted region. $W$ is the scaling factor vector for $K$, and the different gray levels of $W$ represent the magnitudes of the scaling factor for $K$.}
    \label{maskDrivenSelfAttention}
\end{figure}

In the video outpainting task, the given region contains sufficient and complete information, whereas the outpainted region lacks information. In the self-attention layer, the video is divided into patches in the latent space. After injecting the masked video condition information, the patches corresponding to the given region contain sufficient information, whereas the patches corresponding to the outpainted region contain almost no information at the early stages of denoising. Intuitively, attending to the patches of the given region in self-attention helps acquire more useful information, while attending to the patches of the outpainted region may lead to the acquisition of incorrect information.

To better adapt to the video outpainting task, we modify the original self-attention mechanism, resulting in mask-driven self-attention, as illustrated in Figure~\ref{maskDrivenSelfAttention}. The specific formulation is as follows:
\begin{equation}
    \text{Attn}(Q, K, V) = \text{softmax}\left(\frac{Q(K \cdot (1+\gamma \mathcal{F}_s(m)))^T}{\sqrt{d_k}}\right)V,
\end{equation}
where $Q$, $K$, and $V$ are the query, key, and value matrices, respectively, $d_k$ is the channel dimension of the key, $\gamma$ is a hyperparameter, and $\mathcal{F}_s(\cdot)$ is the scaling module composed of linear layers and activation functions with output values constrained to the range $[-1, 1]$. As shown in Figure~\ref{maskDrivenSelfAttention}, the green embeddings represent patches corresponding to the given region, while the gray embeddings represent patches corresponding to the outpainted region. Mask-driven self-attention adjusts the attention weights by scaling the key embeddings based on the mask condition $m$ along with learnable parameters. This enables the model to flexibly allocate attention between the given and outpainted regions, enhancing its adaptability to the video outpainting task.

\subsection{Training Strategy}
During the training process, the attention layers and normalization layers in the DiT blocks, along with the light-weight module in the efficient video control branch, are jointly trained using both a diffusion loss and a latent alignment loss.

\noindent \textbf{Diffusion Loss.}
The training process of our OutDreamer is similar to the video diffusion models aforementioned in Section~\ref{Sec:Preliminaries}. Considering the video condition $Z_{\text{masked}}$, mask condition $m$, and the noisy latent input $Z_t$, the simplified objective of the diffusion reversed process can be calculated by:
\begin{equation}
    \mathcal{L}_{\epsilon} = \mathbb{E}_{\epsilon\sim\mathcal{N}(\mathbf{0},\mathbf{I})} \Big[\vert\vert\epsilon - \epsilon_{\theta}(Z_t, t, Z_{\text{masked}}, m, c_{\text{text}})\vert\vert^2 \Big],
\end{equation}
where $c_{\text{text}}$ is the text condition.


\noindent \textbf{Latent Alignment Loss.}
To enhance overall generation consistency, we compute the mean and variance of each frame for both the predicted latent $\hat{Z}_0$ and the ground truth latents $Z_0$, applying L1 loss to align their mean and variance. The latent alignment loss function is defined as:
\begin{equation}
    \mathcal{L}_{\text{latent}} = \mathbb{E} \left[ | \mu(\hat{Z}_0) - \mu(Z_0) | + | \sigma(\hat{Z}_0) - \sigma(Z_0) | \right],
\end{equation}
where $\mu(\cdot)$ and $\sigma(\cdot)$ represent the mean set and variance set, respectively, formed by calculating the mean and variance of each frame of the input. 

The total loss function is given by:
\begin{equation}
    \mathcal{L} = \mathcal{L}_{\epsilon} + g_{t} \beta \mathcal{L}_{latent},
\end{equation}
where $g_{t} = 1$ when $t < T_{\text{latent}}$; otherwise, $g_{t} = 0$. Here, $\beta$ and $T_{\text{latent}}$ are the hyperparameters. 
Specifically, $T_{\text{latent}}$ denotes the maximum time step for activating the latent alignment loss, indicating that the latent alignment loss operates only during the later stages of the denoising process, when the model has already established the overall structure and requires refinement of specific details.


\subsection{Long Video Generation}
\label{Sec:LongVideoGeneration}
\begin{figure}[t]
    \centering
    \includegraphics[width=\linewidth]{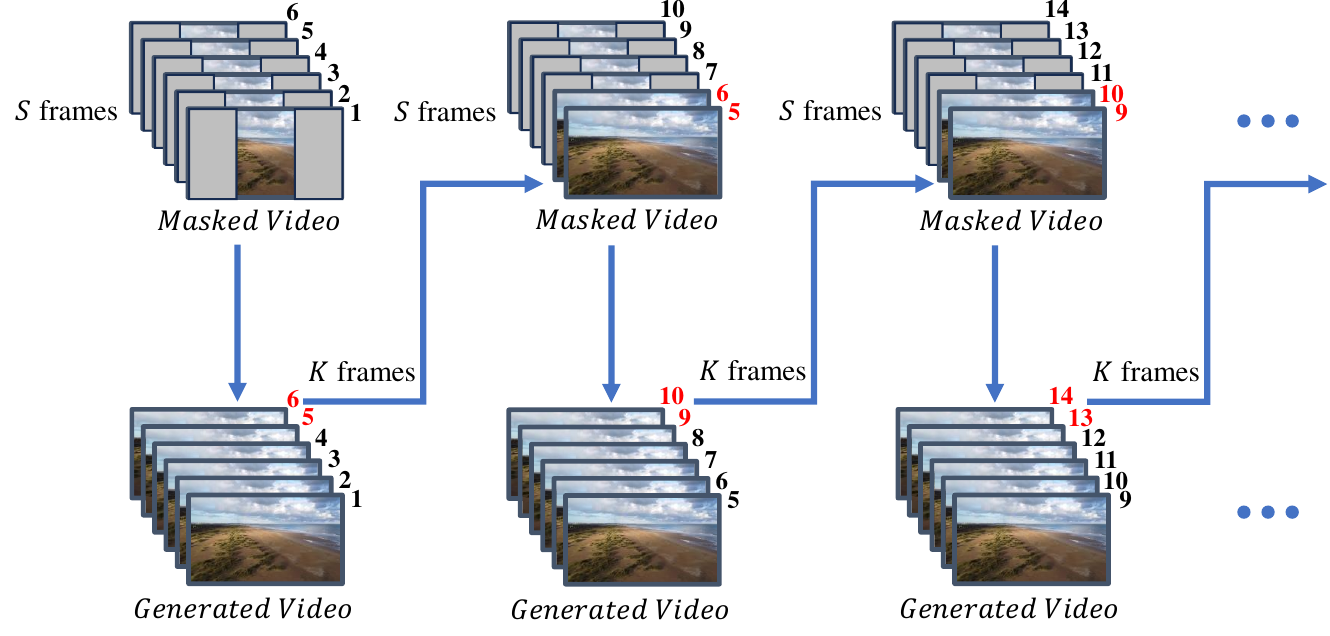}
    \caption{\textbf{Iterative long video generation.} Starting from the second video clip, each clip with total $S$ frames utilizes the last $K$ frames of the previously generated clip as condition information. This example shows the case for $S=6$ and $K=2$.}
    \label{longVideoGeneration}
\end{figure}

For short video outpainting, the aforementioned method is sufficient. However, due to the GPU memory constraint, it is difficult for the DiT to accommodate a long video. To handle this problem, as illustrated in Figure~\ref{longVideoGeneration}, we present an iterative generation approach that utilizes the advanced version of the efficient video control branch (upper-right part in Figure~\ref{overview}) along with the cross-video-clip refiner (in the lower part in Figure~\ref{overview}) to complete the long video outpainting task. The mask condition $m$ injected into the mask-driven self-attention is also replaced with the updated mask condition $\overline{m}$ (in the lower-right part in Figure~\ref{overview}).
Formally, given an input long video $X_{l} \in \mathbb{R}^{H' \times W' \times S_{l} \times 3}$ and its corresponding outpainting masks $M_{l} \in \mathbb{R}^{H \times W \times S_{l} \times 1}$, our objective is to generate an extended long video $\hat{X_{l}} \in \mathbb{R}^{H \times W \times S_{l} \times 3}$, where $H' < H$ and $W' < W$.

\subsubsection{Iterative Generation} 
As shown in Figure~\ref{longVideoGeneration}, we divide the long video into multiple overlapping clips, each containing a total of $S$ frames, with its last $K$ frames overlapping with the subsequent clip. The first clip is generated directly by the basic version of the efficient video control branch and the conditional outpainting branch. Starting from the second clip, each clip is generated using the advanced version of the control branch, which leverages the last $K$ frames from the previously outpainted clip as conditioning information. Specifically, the input masked video $\overline{X}_{masked}$ contains $K$ already outpainted frames from the previously clip and $S-K$ frames that need outpainting. The corresponding masks, $\overline{M}$, include $K$ masks with all values set to $1$ and $S-K$ normal outpainting masks. Similarly to the previously mentioned processing method, $\overline{Z}_{masked}$ is derived from the masked video $\overline{X}_{masked}$ using the VAE encoder $\mathcal E$, while $\overline{m}$ is obtained from the corresponding masks $\overline{M}$ through the downsampling function. Additionally, we employ the cross-video-clip refiner module to refine these clips for maintaining the temporal and color consistency across the entire long video.



\subsubsection{Cross-Video-Clip Refiner}

Iterative generation of a long video sequence often leads to error accumulation and artifacts over time. To address this, we use this refiner to improve the generated video quality. 

Let a generated video clip sequence be $V=\left\{ \{x^i_1, x^i_2, \dots, x^i_S \} \mid i=1,2,\dots, I \right\}$, where $I$ is the total number of video clips, $S$ is the number of frames per clip, and $x_j^i$ is the $j$-th frame of the $i$-th clip.
As shown in Figure~\ref{longVideoGeneration}, there are $K$ overlapping frames between every two adjacent clips. This means that the last $K$ frames $x^{i-1}_{S-K+1:S}$ of the $(i-1)$-th clip should match the first $K$ frames $x^{i}_{1:K}$ of the $i$-th clip.\footnote{Here, $x^{i-1}_{S-K+1:S}$=$\{x^{i-1}_j\}_{j=S-K+1}^S$ represents the last $K$ frames of the $(i-1)$-th clip, and $x^{i}_{1:K}$=$\{x^{i}_j\}_{j=1}^K$ represents the first $K$ frames of the $i$-th clip.} 

The cross-video-clip refiner consists of two main components: mean-variance alignment and histogram matching. 
The mean-variance alignment aims to adjust the mean and variance of $x^{i}_{1:K}$ to match those of $x^{i-1}_{S-K+1:S}$. This process can be viewed as scaling and shifting operations, and the alignment applies the same scaling and shifting operations to the entire clip $x^{i}_{1:S}$. 

The histogram matching aims to adjust the histogram of $x^{i}_{1:K}$ to match that of $x^{i-1}_{S-K+1:S}$. Since all values range from 0 to 255, histogram matching can be viewed as a mapping operation that transforms the values within the range of 0 to 255 to another range of 0 to 255. The histogram matching applies the same mapping operation to the entire clip $x^{i}_{1:S}$. The algorithm of the refinement is provided in the supplementary material.

\section{Experiments}

\subsection{Experimental Setup}

\begin{figure*}[ht]
    \centering
    \includegraphics[width=\linewidth]{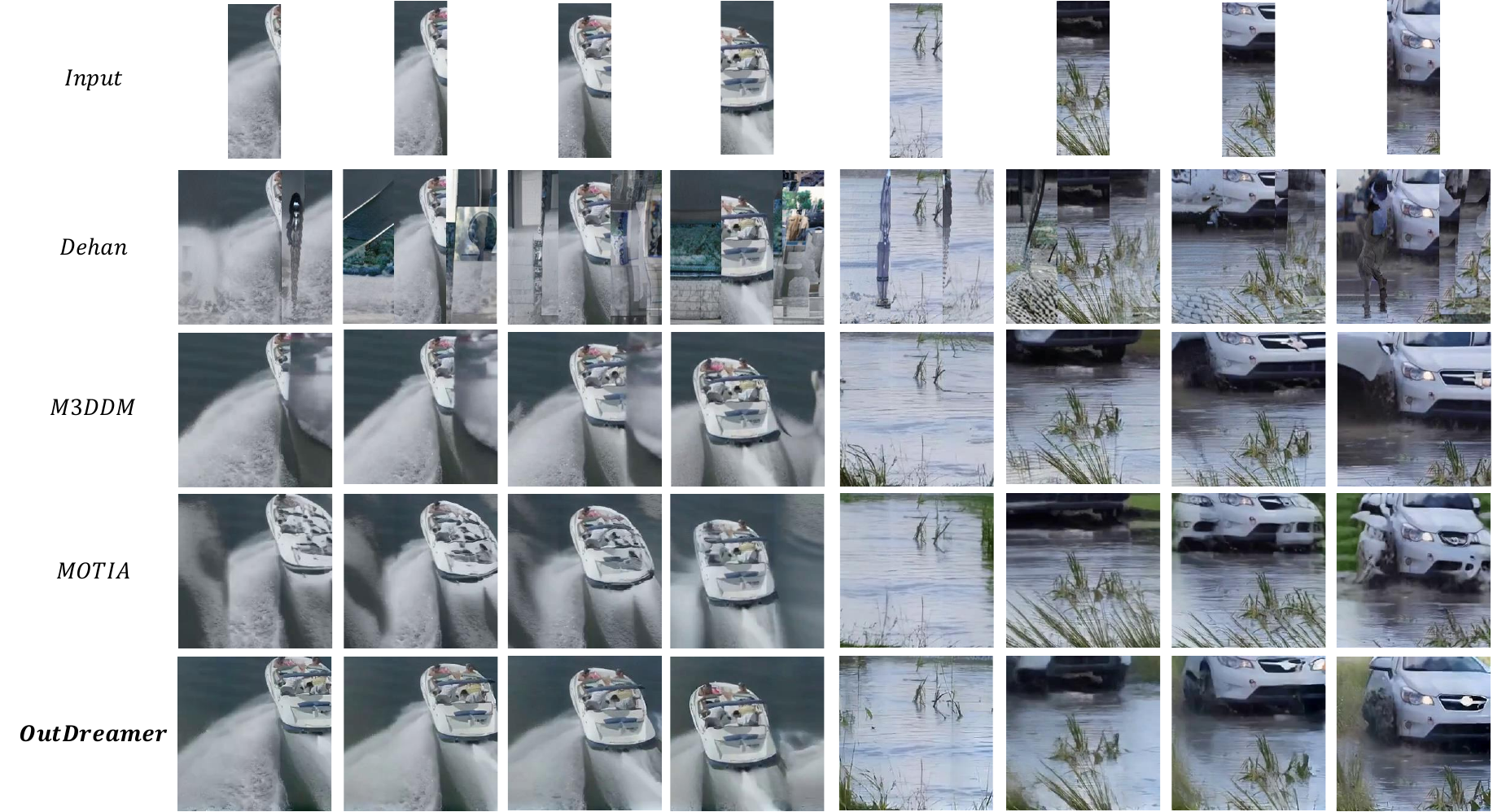}
    \caption{\textbf{Qualitative comparison of video outpainting methods.} The mask ratio is 0.66. We compare our method with Dehan, M3DDM, and MOTIA. SDM is not included since its code is not released.}
    \label{qualitativeResults}
\end{figure*}

\begin{table*}[t]
    \centering
    \begin{tabular}{l|cccc|cccc}
        \hline
        \multirow{2}{*}{Zero-shot method} & \multicolumn{4}{c}{DAVIS dataset} & \multicolumn{4}{c}{YouTube-VOS dataset} \\ \cline{2-9} 
        & SSIM$\uparrow$ & PSNR$\uparrow$ & LPIPS$\downarrow$ & FVD$\downarrow$ & SSIM$\uparrow$ & PSNR$\uparrow$ & LPIPS$\downarrow$ & FVD$\downarrow$   \\ \hline
        Dehan     & 0.6272        & 17.96       & 0.2331       & 363.1       & 0.7195        & 18.25       & 0.2278       & 149.7       \\     \hline
        SDM       & 0.7078        & 20.02       & 0.2165       & 334.6       & 0.7277        & 19.91       & 0.2001       & 94.81       \\     \hline
        M3DDM     & 0.7082        & 20.26       & 0.2026       & 300.0       & 0.7312        & 20.20       & 0.1854       & 66.62       \\     \hline
        OutDreamer      & \textbf{0.7572}          & \textbf{20.30}          & \textbf{0.1742}          & \textbf{268.9}         & \textbf{0.7644}          & \textbf{20.21}          & \textbf{0.1827}          & \textbf{56.02}         \\ \hline
    \end{tabular}
    \caption{\textbf{Quantitative evaluation of zero-shot video outpainting methods on the DAVIS and YouTube-VOS datasets.}}
    \label{tab:Quantitative-Results}
\end{table*}

\subsubsection{Baseline Methods}  
We compare our OutDreamer with three state-of-the-art zero-shot video outpainting methods:
(1) Dehan~\cite{Dehan} utilizes video object segmentation and inpainting techniques to estimate the background, and combines optical flow with image shifting to generate the final outpainted video.
(2) SDM~\cite{M3DDM} uses the first and last frames as conditioning frames that are concatenated with the context video clip in the input layer, without employing mask modeling, and then feeds the combined input into a denoising 3D UNet.
(3) M3DDM~\cite{M3DDM} employs a masked 3D diffusion model, and uses a bidirectional learning approach with mask modeling to guide the generation process with global frames as prompts. 

We also compare our zero-shot OutDreamer with a recent one-shot learning video outpainting method MOTIA~\cite{Be-your-outpainter}. MOTIA fine-tunes a pre-trained model for each specific input video. It consists of two main stages: input-specific adaptation and pattern-aware outpainting.

It is important to note that our approach is designed for zero-shot video outpainting, meaning it can generate missing content directly from the input video without requiring any fine-tuning for the specific video. In contrast, one-shot learning methods like MOTIA require fine-tuning on each individual video before performing outpainting.

\subsubsection{Benchmarks and Evaluation Metrics}  
Following previous works, we evaluate the quantitative performance of OutDreamer on the DAVIS and YouTube-VOS datasets. Horizontal outpainting is performed on the videos using mask ratios of both 0.25 and 0.66, with average scores across these two ratios reported.

The following evaluation metrics are used to assess performance: 
(1) PSNR (Peak Signal-to-Noise Ratio), 
(2) SSIM (Structural Similarity Index), 
(3) LPIPS (Learned Perceptual Image Patch Similarity), 
and (4) FVD (Fréchet Video Distance). 
Following previous work, for PSNR, SSIM, and FVD, the generated results are converted to video frames with pixel values in the range [0, 1]. For LPIPS, we use the range [-1, 1]. For FVD, we uniformly sample 16 frames from each video for evaluation. 

\subsubsection{Implementation Details}  

\begin{figure*}[ht]
    \centering
    \includegraphics[width=\linewidth]{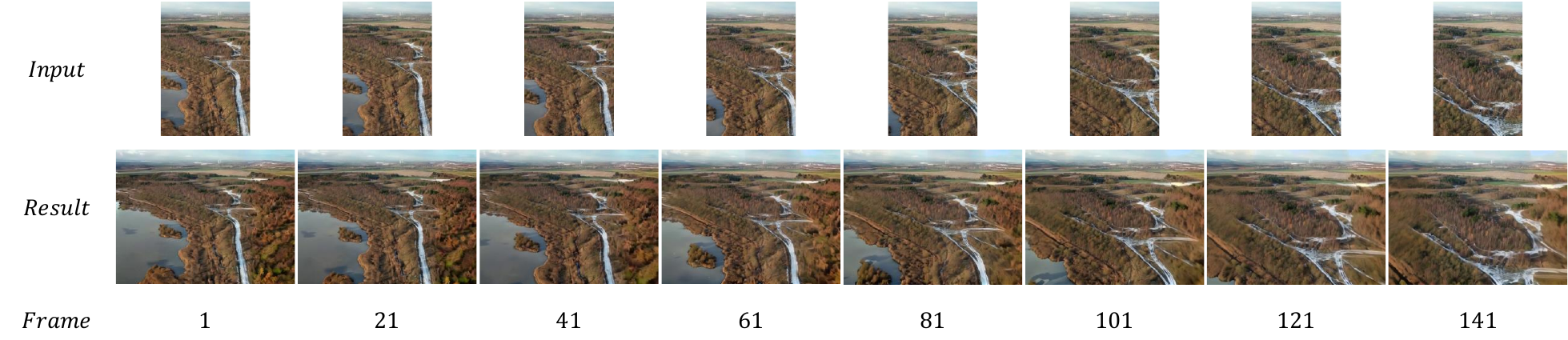}
    \caption{\textbf{A long video generation example.} The mask ratio is 0.5.}
    \label{longVideoExample}
\end{figure*}

Our method is based on the Open-Sora-Plan~\cite{Open-Sora-Plan} T2V (Text-to-Video) model, which is pre-trained to generate 29 frames of 640 $\times$ 480 videos from text captions. During the sampling process, we set the classifier-free guidance (CFG) scale to 3, and the number of diffusion steps to 100. 
For a fair comparison, we use a rough caption for each video automatically generated by ShareGPT4V~\cite{chen2024sharegpt4video, chen2023sharegpt4v, chen2024we} along with a random frame from this video.
As is common in video outpainting, we blend the input video into the generated video to obtain the final result.
In long video generation experiments, the number of condition frames from the previous clip, $K$, is set to 3, while the number of frames per clip, $S$, remains 29. 


\subsection{Qualitative Results}

Figure~\ref{qualitativeResults} illustrates the qualitative results of our OutDreamer compared to other zero-shot and one-shot learning baseline methods. 
Dehan exhibits noticeable artifacts and visual inconsistencies. M3DDM fails to correctly outpaint the ship in the right part of the frame. MOTIA successfully completes the outpainting of the ship, but has issues with repeating and misaligned elements when outpainting the car. In contrast, our method generates realistic and temporally consistent results when compared to both the zero-shot and one-shot learning baselines.

\subsection{Quantitative Results}

Table~\ref{tab:Quantitative-Results} presents the evaluation results of OutDreamer compared with other zero-shot methods on the DAVIS and YouTube-VOS datasets. Our method outperforms them in all the scores. These results demonstrate that OutDreamer produces video outpainting with better quality and temporal consistency.

Table~\ref{tab:zero-shot-vs-one-shot} compares our zero-shot method with the one-shot learning method MOTIA on the YouTube-VOS dataset. OutDreamer achieves comparable performance to MOTIA in terms of the PSNR, SSIM, LPIPS, and FVD scores. These results indicate that video outpainting methods can be both efficient and high-quality in a zero-shot setting, without the need for fine-tuning on each input video as required by one-shot learning methods.

\begin{figure}[ht]
    \centering
    \includegraphics[width=\linewidth]{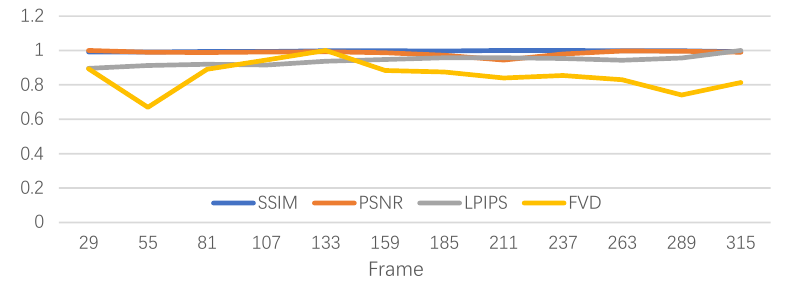}
    \caption{\textbf{Quantitative evaluation of long video generation.} For each metric score (PSNR, SSIM, LPIPS, and FVD), we perform maximum normalization with proportional scaling, making the results more intuitive.
    }
    \label{longVideoChart}
\end{figure}

\subsection{Long Video Generation}

We assess the performance of our iterative generation method for long video generation. Figure~\ref{longVideoExample} displays an example generated by our method. The results demonstrate that, with the support of our previous condition frames and the cross-video-clip refiner, temporal consistency is well-maintained even in long video sequences.


\begin{table}[t]
    \centering
    \begin{tabular}{c|cccc}
        \hline
        Method & SSIM$\uparrow$ & PSNR$\uparrow$ & LPIPS$\downarrow$ & FVD$\downarrow$   \\ \hline
        One-shot & \multirow{2}{*}{0.7636}        & \multirow{2}{*}{20.25}       & \multirow{2}{*}{0.1727}       & \multirow{2}{*}{58.99}       \\ 
        MOTIA &        &        &        &        \\ \hline
        Zero-shot & \multirow{2}{*}{0.7644}        & \multirow{2}{*}{20.21}       & \multirow{2}{*}{0.1827}       & \multirow{2}{*}{56.02}       \\ 
        OutDreamer &        &        &        &        \\ \hline
    \end{tabular}
    \caption{\textbf{Comparison with one-shot learning method MOTIA on the YouTube-VOS dataset.}}
    \label{tab:zero-shot-vs-one-shot}
\end{table}

For further evaluation, we construct a simple long-video dataset of 20 long videos sourced from the Pexels website\footnote{https://www.pexels.com}, generate videos with varying frame lengths using mask ratio of 0.5 and calculate the PSNR, SSIM, LPIPS, and FVD scores for the last clip in each video. Figure~\ref{longVideoChart} presents the results as the frame length increases from 29 to 315, indicating that our iterative method preserves high quality and temporal consistency, even for long video generation.

\subsection{Ablation Study}

To evaluate the effectiveness of our latent alignment loss and mask-driven self-attention, we conduct an ablation study on the YouTube-VOS dataset. We train three models—one baseline, one with only the latent alignment loss, and one with both the latent alignment loss and the mask-driven self-attention—and compare their performances. Table~\ref{tab:latent-loss-ablation} shows the quantitative comparison, indicating that the latent alignment loss and the mask-driven self-attention improve the quality of the generated videos, with the model incorporating both achieving the best performance.





Additionally, we conduct an ablation study to evaluate the effectiveness of the cross-video-clip refiner. We compare the long video outpainting results with and without the refiner, and evaluate the results using PSNR, SSIM, LPIPS, and FVD metrics on our long-video dataset. Each video is outpainted to 315 frames with a mask ratio of 0.5. As shown in Table~\ref{tab:refiner-ablation}, the cross-video-clip refiner improves the quality of the outpainted videos on all metrics. The supplementary material provides an ablation study verifying the effectiveness of the downsampling function for the mask condition.


\begin{table}[t]
    \centering
    \begin{tabular}{l|cccc}
        \hline
        Method & SSIM$\uparrow$ & PSNR$\uparrow$ & LPIPS$\downarrow$ & FVD$\downarrow$   \\ \hline
        / & 0.7602        & 19.66       & 0.2002       & 60.84       \\ \hline
        LAL  & 0.7620        & 19.99       & 0.1846       & 58.82       \\ \hline
        LAL+MSA  & 0.7644        & 20.21       & 0.1827       & 56.02       \\ \hline
    \end{tabular}
    \caption{\textbf{Ablation study of the latent alignment loss and mask-driven self-attention on the YouTube-VOS dataset.} LAL: latent alignment loss. MSA: mask-driven self-attention.}
    \label{tab:latent-loss-ablation}
\end{table}

\begin{table}[t]
    \centering
    \begin{tabular}{l|cccc}
        \hline
        Method & SSIM$\uparrow$ & PSNR$\uparrow$ & LPIPS$\downarrow$ & FVD$\downarrow$   \\ \hline
        w/o refiner & 0.7219        & 17.31       & 0.2263       & 276.9       \\ \hline
        w/ refiner  & 0.7303        & 17.37       & 0.2256       & 256.2       \\ \hline
    \end{tabular}
    \caption{\textbf{Ablation study of the cross-video-clip refiner.}}
    \label{tab:refiner-ablation}
\end{table}
\section{Conclusion}

In this paper, we propose a novel zero-shot video outpainting method based on the DiT architecture with efficient condition injection. Our approach injects condition information at the early stages of the network, applies mask-driven self-attention to better adapt to the outpainting task, introduces a latent alignment loss function to mitigate overall inconsistency, and employs an iterative generation strategy to enable the generation of long video sequences while preserving temporal and color consistency across frames. Experimental results demonstrate that our method is capable of generating high-quality outpainted videos. 



{
    \small
    \bibliographystyle{ieeenat_fullname}
    \bibliography{main}
}

\newcounter{mysec}[section]
\renewcommand{\themysec}{\Roman{mysec}}
\newcommand{\mysec}[1]{
  \refstepcounter{mysec}
  \subsection*{\themysec. #1}
}

\clearpage
\mysec{Details of the Latent Alignment Loss}
The latent alignment loss function supervises the overall mean and variance of each generated frame in the latent space, enhancing overall consistency. 
Formally, let $Z_0$ represent the ground truth frames in the latent space. The noisy sample at time $t$ is expressed as:
\begin{equation}
    Z_t = \sqrt{\bar{\alpha}_t} Z_0 + \sqrt{1-\bar{\alpha}_t} \epsilon,
\end{equation}
where $\epsilon \sim \mathcal{N}(\mathbf{0}, \mathbf{I})$. Based on the predicted noise $\epsilon_{\theta}(Z_t, t, Z_{\text{masked}}, m, c_{\text{text}})$, the estimated latent frames $\hat{Z}_0$ is computed as:
\begin{equation}
    \hat{Z}_0 = \frac{Z_t - \sqrt{1-\bar{\alpha}_t} \epsilon_{\theta}(Z_t, t, Z_{\text{masked}}, m, c_{\text{text}})}{\sqrt{\bar{\alpha}_t}}.
\end{equation}
We compute the L1 loss of the means and variances between the predicted $\hat{Z}_0$ and the ground truth $Z_0$.

\mysec{Algorithm of Cross-Video-Clip Refiner}
We present the process of the cross-video-clip refiner using Python pseudocode, as shown in Algorithm~\ref{alg:refiner}. In the refiner, the mean-variance alignment and the histogram matching are performed independently for each channel of the RGB color space.

\definecolor{keywordcolor}{RGB}{0,0,255}     
\definecolor{stringcolor}{RGB}{163,21,21}   
\definecolor{commentcolor}{RGB}{0,128,0}    
\definecolor{backgroundcolor}{RGB}{245,245,245} 

\lstset{
    language=Python,                     
    basicstyle=\ttfamily\small,          
    keywordstyle=\color{keywordcolor}\bfseries, 
    stringstyle=\color{stringcolor},     
    commentstyle=\color{commentcolor}\itshape,  
    backgroundcolor=\color{backgroundcolor},    
    frame=single,                        
    numbers=left,                        
    numberstyle=\tiny\color{gray},       
    breaklines=true,                     
    tabsize=4                            
}

\begin{algorithm*}
\caption{\textbf{Python Pseudocode of the Cross-Video-Clip Refiner}. The function \textbf{mean\_variance\_alignment()} implements the mean-variance alignment, while \textbf{histogram\_matching()} implements the histogram matching. In addition, \textbf{source\_video} represents the first $k$ frames from the current clip, \textbf{template\_video} represents the last $k$ frames from the previously outpainted clip, and \textbf{target\_video} represents the current clip. This Python code is a simplified version of the actual implementation, focusing only on the core part of the refiner to enhance readability.}
\label{alg:refiner}
\begin{lstlisting}
import cv2
import numpy as np
def video_stats(video):
    r, g, b = video[..., 0], video[..., 1], video[..., 2]
    rMean, rStd = r.mean(), r.std()
    gMean, gStd = g.mean(), g.std()
    bMean, bStd = b.mean(), b.std()
    return rMean, rStd, gMean, gStd, bMean, bStd
def mean_variance_alignment(source_video: np.ndarray, template_video: np.ndarray, target_video: np.ndarray) -> np.ndarray:
    # source_video/template_video: [K, H, W, C]; target_video: [T, H, W, C]
    rMeanSrc, rStdSrc, gMeanSrc, gStdSrc, bMeanSrc, bStdSrc = video_stats(source_video)
    rMeanTmpl, rStdTmpl, gMeanTmpl, gStdTmpl, bMeanTmpl, bStdTmpl = video_stats(template_video)
    ans = np.zeros_like(target_video)
    for i in range(len(target_video)):
        r, g, b = cv2.split(target_video[i])
        r -= rMeanSrc; g -= gMeanSrc; b -= bMeanSrc
        r = (rStdTmpl / rStdSrc) * r
        g = (gStdTmpl / gStdSrc) * g
        b = (bStdTmpl / bStdSrc) * b
        r += rMeanTmpl; g += gMeanTmpl; b += bMeanTmpl
        r = np.clip(r, 0, 255); g = np.clip(g, 0, 255); b = np.clip(b, 0, 255)
        ans[i] = cv2.merge([r, g, b])
    return ans
def histogram_matching(source_video: np.ndarray, template_video: np.ndarray, target_video: np.ndarray) -> np.ndarray:
    # source_video/template_video: [K, H, W, C]; target_video: [T, H, W, C]
    ans = np.zeros_like(target_video)
    for i in range(3):
        src_hist = cv2.calcHist([source_video[..., i].reshape(-1)], [0], None, [256], (0, 256), accumulate=False)[:, 0]
        tmpl_hist = cv2.calcHist([template_video[..., i].reshape(-1)], [0], None, [256], (0, 256), accumulate=False)[:, 0]
        src_quantiles = np.cumsum(src_hist) / sum(src_hist)
        tmpl_quantiles = np.cumsum(tmpl_hist) / sum(tmpl_hist)
        tmpl_values = np.arange(0, 256)
        interp_a_values = np.interp(src_quantiles, tmpl_quantiles, tmpl_values)
        for j in range(len(target_video)):
            tar_lookup = target_video[j][..., i].reshape(-1)
            interp = interp_a_values[tar_lookup]
            ans[j][..., i] = interp.reshape(target_video[j][..., i].shape)
    return ans
    \end{lstlisting}
\end{algorithm*}

\mysec{More Implementation Details}

The hyperparameter $\beta$ in the latent alignment loss is set to 0.02, and the maximum time step $T_{\text{latent}}$ for activating the latent alignment loss is set to 200.

During training, a mask ratio between 0.1 and 0.8 is randomly chosen for each video, followed by a random selection of horizontal or vertical masking. Outpainting for horizontal masking generates the left and right parts of video frames, with the mask ratio on each side also chosen randomly. Vertical masking follows the same rule on the top and bottom of video frames. 

For evaluation, the generated frames are resized to 256 $\times$ 256 to ensure a fair comparison with other methods. The LPIPS score is calculated using AlexNet, while the FVD score is computed using VideoGPT. For the FVD evaluation, as our model generates 29 frames per clip, we uniformly sample 29 frames from each video for outpainting and then uniformly sample 16 frames from the generated video for evaluation. If a video contains fewer than 29 frames, a video interpolation model\footnote{https://github.com/hzwer/ECCV2022-RIFE} is used to interpolate the frames to 29.


\mysec{Additional Ablation Study}



\begin{table}[t]
    \centering
    \begin{tabular}{l|cccc}
        \hline
        Method & SSIM$\uparrow$ & PSNR$\uparrow$ & LPIPS$\downarrow$ & FVD$\downarrow$   \\ \hline
        VAE masks       & 0.7661       & 20.23       & 0.1845    & 60.39   \\ \hline
        downsampling masks &   0.7644        & 20.21       & 0.1827       & 56.02       \\ \hline
    \end{tabular}
    \caption{\textbf{Ablation study of the latent masks on the YouTube-VOS dataset.}}
    \label{tab:mask-ablation}
\end{table}



To verify the effectiveness of the simple downsampling function for the mask condition, we compare two different methods for obtaining the latent masks $m$ from the pixel masks $M$. The first method uses a simple downsampling function, while the second method involves repeating $M$ along the channel dimension and then applying the VAE encoder $\mathcal E$ to obtain the latent masks $m$. Similarly, $\overline{m}$ is derived from the pixel masks $\overline{M}$ using the same approach. The quantitative results of this ablation study on the YouTube-VOS dataset are presented in Table~\supptabref{tab:mask-ablation}. The findings indicate that the simple downsampling function is already sufficiently effective for the mask condition, and using a more complex VAE for prior mask processing does not significantly improve information utilization.

\mysec{Additional Results}

\begin{figure*}[ht]
    \centering
    \includegraphics[width=\linewidth]{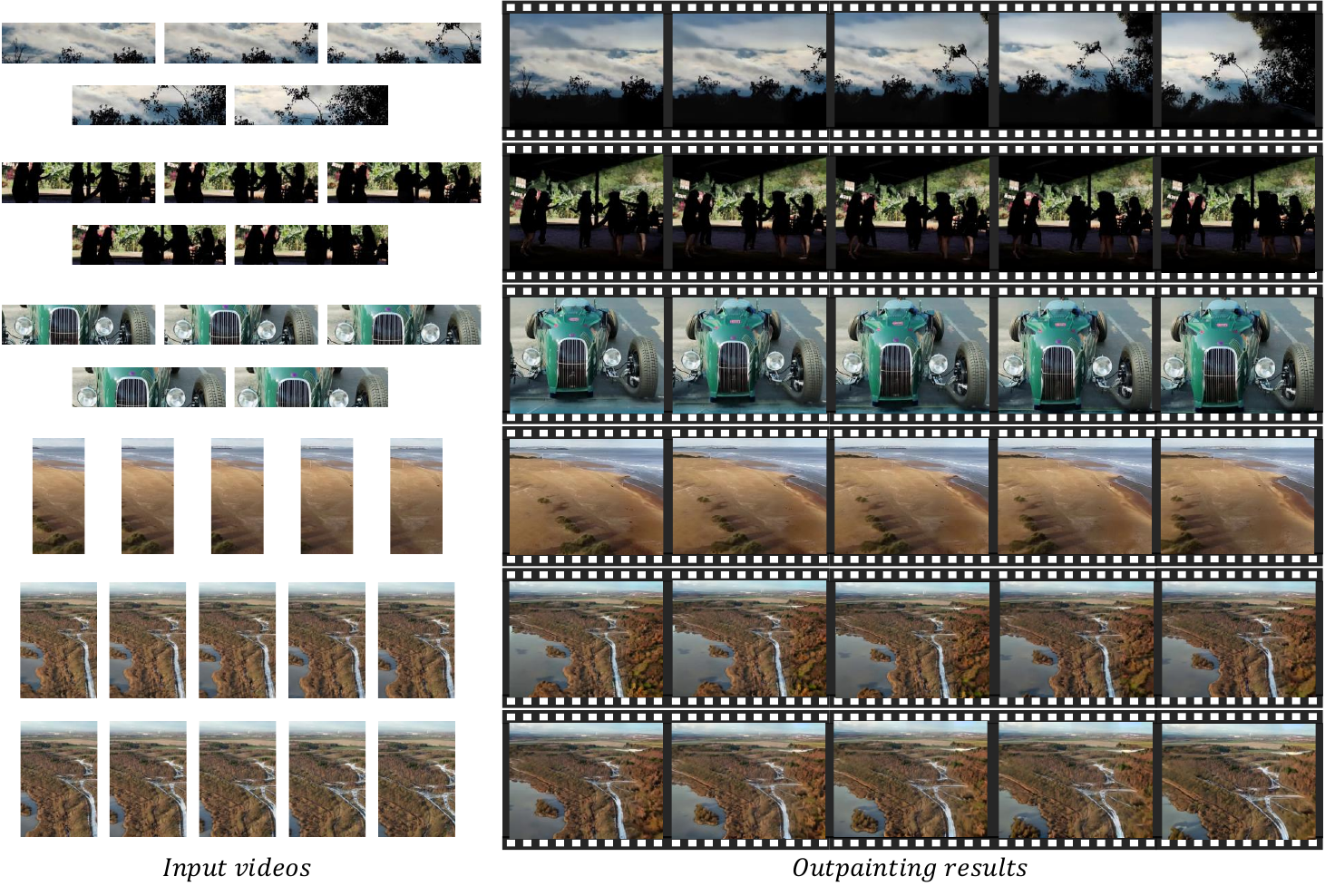}
    \caption{\textbf{Additional outpainting results.} Left: input videos. Right: outpainting results generated by OutDreamer.}
    \label{additional_results}
\end{figure*}

We provide additional outpainting results generated by OutDreamer in Figure~\suppfigref{additional_results}. 

\mysec{Demo Videos}

We provide demo videos outpainted by OutDreamer in the supplementary material. These include high-quality outpainting results and comparisons with baseline methods. 

\end{document}